%% file: acmart.tex
\documentclass[sigconf]{acmart}
\renewcommand\footnotetextcopyrightpermission[1]{} 
\usepackage[titlenumbered,ruled]{algorithm2e}
\usepackage{enumitem}
\usepackage{outlines}
\usepackage{xspace}
\usepackage{booktabs}
\usepackage{wasysym}
\usepackage{tabularx}
\usepackage{tabu}
\usepackage{makecell, rotating} 
\usepackage{subcaption}
\usepackage{caption, booktabs}     
\usepackage{array}
\usepackage{adjustbox}
\usepackage{setspace}
\usepackage{enumitem}
\usepackage[frozencache=true,cachedir=minted-cache]{minted} 
\usepackage{breqn}

\usepackage{relsize} 
\usepackage{scalerel} 
\usepackage{threeparttable}
\usepackage{appendix}

\usepackage{graphicx}
\usepackage{caption}
\usepackage{subcaption}
\usepackage{adjustbox}
\usepackage{multicol}

\usepackage{gensymb}
\usepackage[skip=2pt]{caption}
\usepackage{tikz} 

\usepackage{lscape}
\newcommand{\PreserveBackslash}[1]{\let\temp=\\#1\let\\=\temp}
\newcolumntype{C}[1]{>{\PreserveBackslash\centering}p{#1}}
\newcolumntype{R}[1]{>{\PreserveBackslash\raggedleft}p{#1}}
\newcolumntype{L}[1]{>{\PreserveBackslash\raggedright}p{#1}}

\def\arrvline{\hfil\kern\arraycolsep\vline\kern-\arraycolsep\hfilneg}

\usepackage{amsmath}

\setcopyright{acmcopyright}

\begin{document}
\title{B2RL: An open-source Dataset for Building \newline Batch Reinforcement Learning}

\author{Hsin-Yu Liu$^{*}$, Xiaohan Fu, Bharathan Balaji$^\dagger$, Rajesh Gupta, Dezhi Hong$^\dagger$}
\affiliation{University of California, San Diego, La Jolla, CA, USA, $^\dagger$Amazon}

\thanks{$^*$Corresponding author.}
\thanks{$^\dagger$Work unrelated to Amazon}

\renewcommand{\shortauthors}{Liu et al.}

\begin{abstract}
    Batch reinforcement learning (BRL) is an emerging research area in the RL community. It learns exclusively from static datasets (i.e. replay buffers) without interaction with the environment. In the offline settings, existing replay experiences are used as prior knowledge for BRL models to find the optimal policy. Thus, generating replay buffers is crucial for BRL model benchmark. In our B2RL (\underline{B}uilding \underline{B}atch \underline{RL}) dataset, we collected real-world data from our building management systems, as well as buffers generated by several behavioral policies in simulation environments. We believe it could help building experts on BRL research. To the best of our knowledge, we are the first to open-source building datasets for the purpose of BRL learning.
\end{abstract}



\copyrightyear{2022}
\acmYear{2022}
\acmPrice{15.00}
\acmDOI{10.1145/3360322.3360868}
\acmISBN{978-1-4503-7005-9/19/11}
\maketitle

\input{1_intro}

\input{2_related}

\input{3_approach_and_results}

\input{4_conclusion_discussion}

\section*{Acknowledgement}
    This work was supported in part by the CONIX Research Center, one of six centers in JUMP, a Semiconductor Research Corporation (SRC) program sponsored by DARPA.       

\bibliography{acmart}
\bibliographystyle{ACM-Reference-Format}

\end{document}

%% file: 1_intro.tex
\vspace{-2mm}
\section{Introduction}
Reinforcement learning (RL) is widely studied in the building research area.
Most studies focus on RL learning in an online paradigm ~\cite{RL_for_control_of_building_HVAC_raman2020reinforcement, RL_building_control_opportunity_wang2020reinforcement, MB2C_ding2020mb2c,DeepComfort_gao2020deepcomfort, BuildSys_2019_1_zhang2019building, Practical_RL_zhang2018practical}, assuming there is a
simulation environment for RL models to interact with during training and 
evaluation stages before real-world deployment. Simulators such as
EnergyPlus~\cite{EnergyPlus_crawley2001energyplus} and 
TRNSYS~\cite{TRNSYS_klein1976university} are used to simulate the
thermal states of a building. However, designing and calibrating such models
for a large building is time-consuming and requires expertise. 

    In real-world scenarios, most large buildings are controlled via building
management systems (BMS), where thermal data can be stored in database. 
With advances in sensing technologies and machine learning, data-driven models
have been more popular in recent research. Batch reinforcement learning, a data-driven
approach that learns only from fixed dataset generated with unknown behavioral policy,
has not been explored widely in the building control community. BRL models are
capable of learning the optimal policy without accurate environment models or
simulation environments as oracles. In our study, we open-source both our dataset
\footnote{\url{https://github.com/HYDesmondLiu/B2RL}}
extracted from real building and the one generated with Sinergym~\cite{Sinergym}, a 
building RL simulation environment which integrates EnegryPlus and BCVTB~\cite{BCVTB_wetter2008building} with OpenAI Gym~\cite{openai_gym_brockman2016openai} interface. Furthermore, we experiment with several state-of-the-art BRL methods. The experimental results could be re-used as benchmarks for algorithm comparison.

%% file: 2_related.tex
\section{Related Work}
    \subsection{Building batch reinforcement learning}
    
    Previously, several studies implement fitter Q-iteration (FQI) and batch Q-learning~\cite{BRL_balancing_vazquez2017balancing, BRL_electricity_cost_ruelens2014demand, BRL_Demand_response_application_ruelens2016residential, BRL_Exergy_yang2015reinforcement}. However, for FQI and Batch Q-learning, they
    are based on pure off-policy algorithms. Fujimoto et
al. [23] show that off-policy methods exacerbate the extrapolation
error in a pure offline setting. These errors are attributed to Q-network
training on historical data but exploratory actions yield
policies which are different from the behavioral ones.
    
    Recently, several studies related to building deep BRL research have emerged. Zhang et al. 
    ~\cite{HVAC_BRL} apply CQL~\cite{CQL} on the CityLearn ~\cite{citylearn} testbed as simulator. Liu et al. ~\cite{ICCPS_BRL} incorporates a Kullback-Leibler term in 
    Q-update to penalize policies that are far from the previous one to improve from 
    state-of-the-art BRL algorithm and deploy in real environments without setting up simulators.
    
    \subsection{Batch reinforcement learning datasets}
    
    To our best knowledge, the only open-source BRL dataset is the D4RL
    dataset~\cite{D4RL}. They have generated various robotic control datasets~\cite{mujoco, Adroit, flow, kitchen, carla}. 
    In our study, we open-source two building datasets, one contains real building buffers extracted from our building database with sensor readings, setpoints control history, and the estimated energy consumption calculated by 
    Zonepac~\cite{zonepac_balaji2013zonepac}. Then, we process them as Markov Decision Process (MDP) tuples. 
    The other one is a set of buffers that contain different qualities generated with simulation environments.

%% file: 3_approach_and_results.tex
\section{Approach and Results}
 
  \subsection{Real building buffers}
    \subsubsection{Data acquisition}
        The real building buffer is extracted from the readings of student labs in one of the school 
        buildings. The amount of datapoints in the buffers ranges from  $170$$\sim$$260K$, depending on the
        number of rooms involved and missing values. We obtain data of an entire year, from the
        beginning of July 2017 to the end of June 2018 for 15 rooms across 3 floors. 
        The RL setup in our experiments is listed as below:
                
        \begin{itemize}[leftmargin=*,noitemsep,topsep=0pt,parsep=0pt,partopsep=0pt]
            \item \textit{State}: Indoor air temperature, actual supply airflow, outside air temperature, and humidity.
            
            \item \textit{Action}: Zone air temperature setpoint and actual supply airflow setpoint. Both are in continuous space and the action spaces are normalized in the range of $[-1,1]$ as a standard RL settings.
            
            \item \textit{Reward}: Our reward function is a linear combination of thermal comfort and energy consumption. The reward function at time step $t$ is:
            \begin{center} 
            \small
                \vspace{-2em}
                \begin{equation} \label{eq:reward_function}
                 R_t = -\alpha|TC_t|-\beta P_t,
                \end{equation}
            \end{center}
            where $\alpha$, $\beta$ are the weights balancing different objectives and
            could be tuned to meet specific goals, $TC_t$ is the thermal comfort index at time $t$, $P_t$ is the HVAC power consumption at time $t$. We compute $P_t$ attributed to a thermal zone using heat transfer equations~\cite{zonepac_balaji2013zonepac}.
            
        \end{itemize}

    \subsubsection{BRL benchmarks}
        \begin{itemize} 
                \item Batch-constrained deep Q-learning (BCQ)~\cite{BCQ_fujimoto2019off}: BCQ is a model-free RL method that mitigates extrapolation errors induced by incorrect value estimation of out-of-distribution actions selected out of existing dataset. 

                \item Bootstrapping Error Accumulation Reduction (BEAR)~\cite{BEAR_kumar2019stabilizing}: BEAR identifies bootstrapping error as a key source of BRL instability. The algorithm mitigates out-of-distribution action selection by searching over the set of policies that is akin to the behavior policy.
                
                \item Pessimistic Q-Learning (PQL)~\cite{PQL_liu2020provably}:
                PQL uses pessimistic value estimates in the low-data regions in the Bellman optimality equation as well as the evaluation back-up. It can yield stronger guarantees when the concentrability assumption does not hold. PQL learns from policies that satisfy a bounded density ratio assumption similar to on-policy policy gradient methods.
        \end{itemize}
        
    \subsubsection{Experiment details}
        Each algorithm is run in one
        room on each floor in the entire week so that outside air temperature
        (OAT) is the same. For instance, in one week we run algorithm A in
        rooms in the same stack on different floors, e.g. 2144, 3144, and
        4144, and at the same time algorithm B runs on 2146, 3146, and 4146. In each room, we train the algorithm for 1,000 time steps,
        which is about one week. We evaluate each algorithm in three different rooms (one room from
        each floor: 2F, 3F, and 4F). These rooms are of roughly the same size and occupancy
        capacity. Each time step is 10 minute due to the data
        writing rate in our BMS. More details of the experiments are described previously in our previous study~\cite{HVAC_BRL}.
        
\begin{figure}[h]
                \begin{center}
                \includegraphics[width=.7\linewidth]{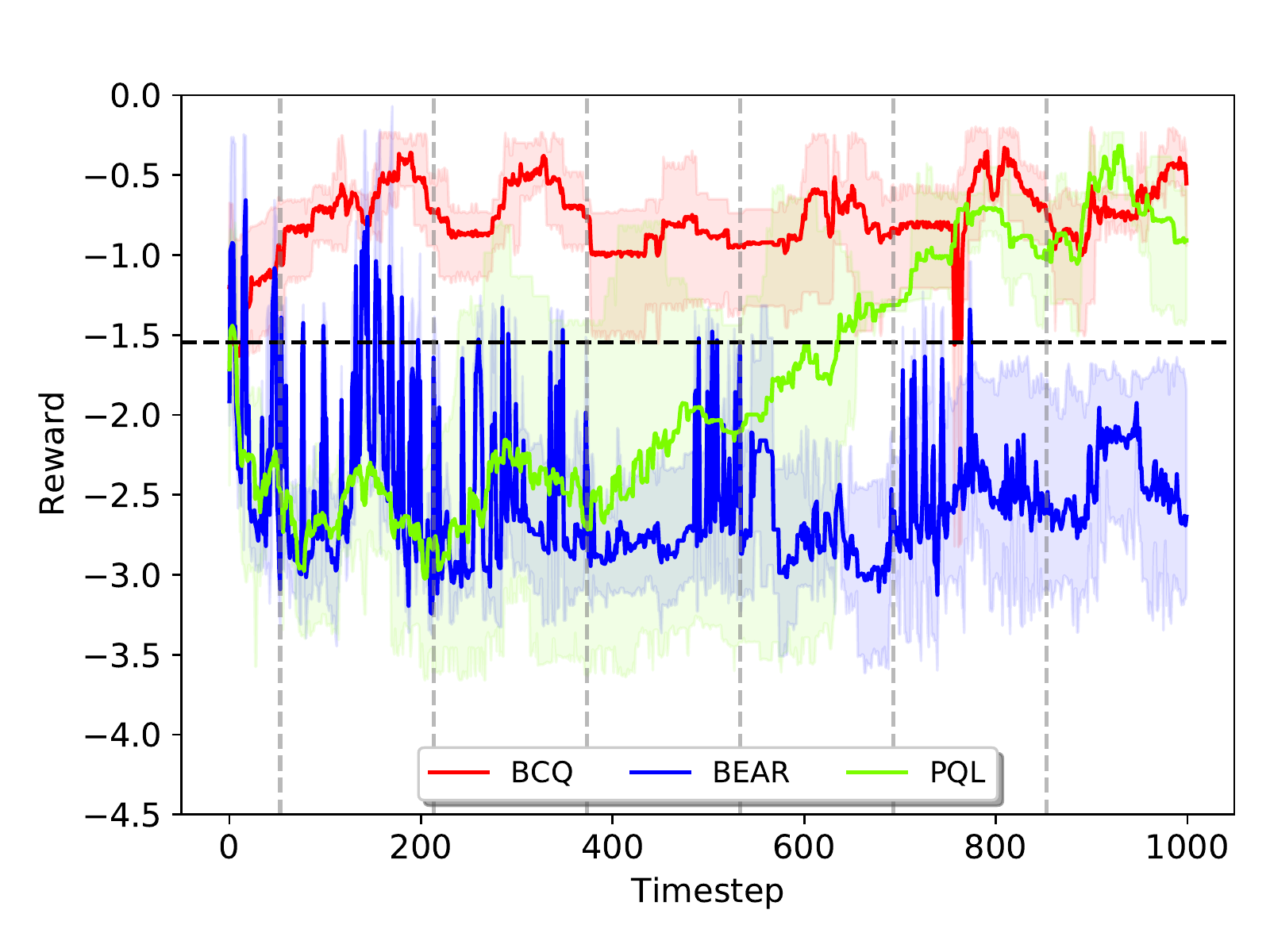}
                \caption{Episode reward comparison in real building}
                \label{fig:real_bldg_lc}
                \vspace{-5mm}
                \end{center}
            \end{figure}

\begin{figure}[h]
                \begin{center}
                \includegraphics[width=0.7\linewidth]{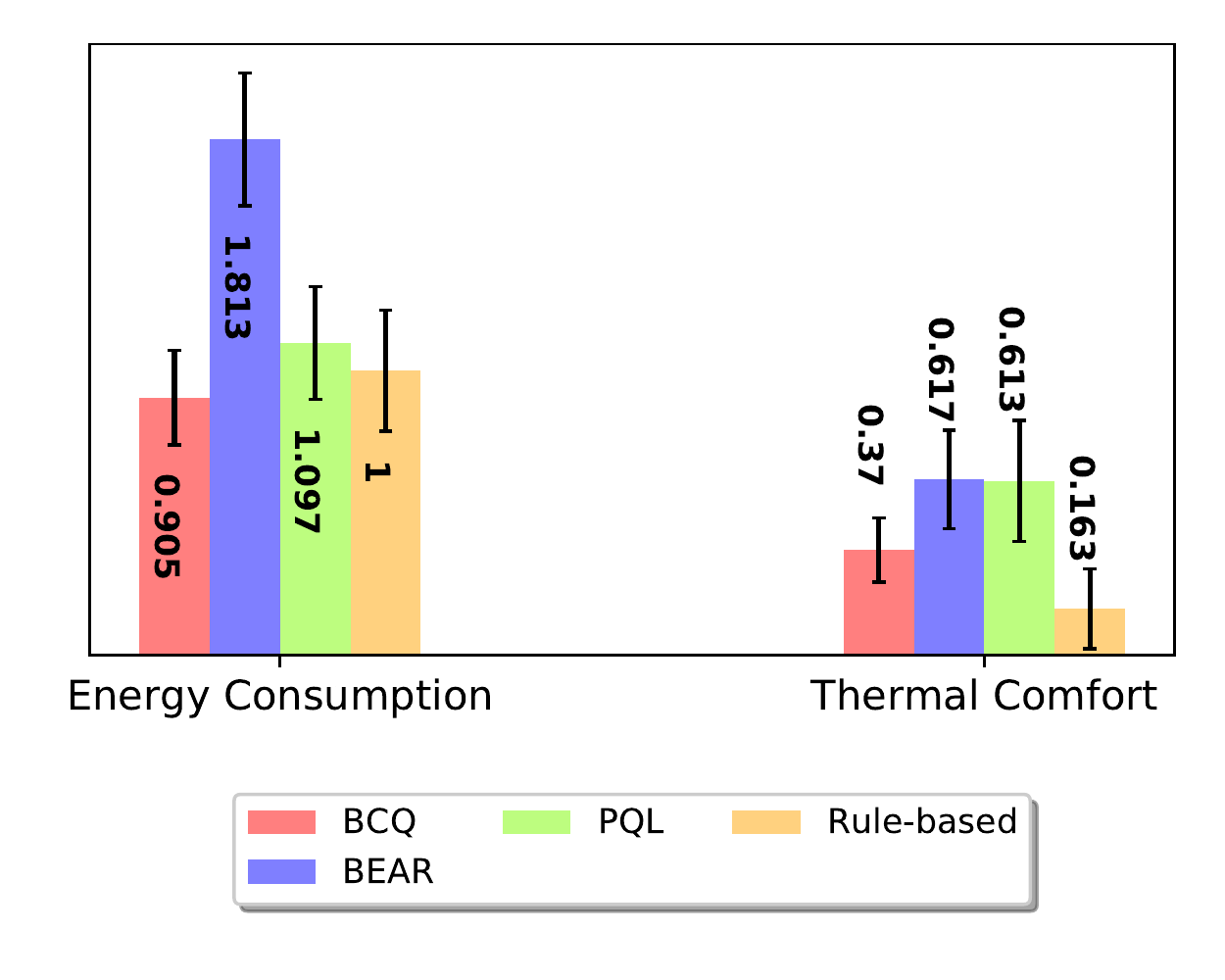}
                \caption{Optimization objectives analysis in real building}
                \label{fig:real_bldg_oo}
                \vspace{-5mm}
            \end{center}
            \end{figure}

        Fig.~\ref{fig:real_bldg_lc} shows the learning curves of each algorithm, where each
        solid line is the average reward of all runs for the same method; semi-transparent
        bands represent the range of all runs for a particular algorithm. And gray dotted
        vertical lines indicate 00:00AM of each day. The horizontal black dotted line
        is the average reward in the buffer. Fig.~\ref{fig:real_bldg_oo} shows the analysis of the optimization objectives in the reward function, for energy consumption, the default control method rule-based contorl (RBC) method is normalized to
        1. And for thermal comfort we are showing absolute averaged values.

        As we need to calculate the thermal comfort level as required by our reward function, we adopt the widely used predicted mean vote (PMV)~\cite{PMV_fanger1970thermal} measure as our thermal comfort index. 
        In this metric, thermal comfort satisfaction ranges from $-3$ (cold) to $3$ (hot), where PMV within the range of $-0.5$ to $0.5$ is considered as thermal comfortable.
        We adopt the ASHRAE RP-884 thermal comfort data set~\cite{ASHRAE_TC_de1998global} and train a simple gradient boosting tree (GBT) model~\cite{LightGBM_ke2017lightgbm} to predict the thermal comfort by taking the current thermal states given by our building system in real-time.

                \begin{figure}[ht]
            \centering
         \includegraphics[width=.9\linewidth]{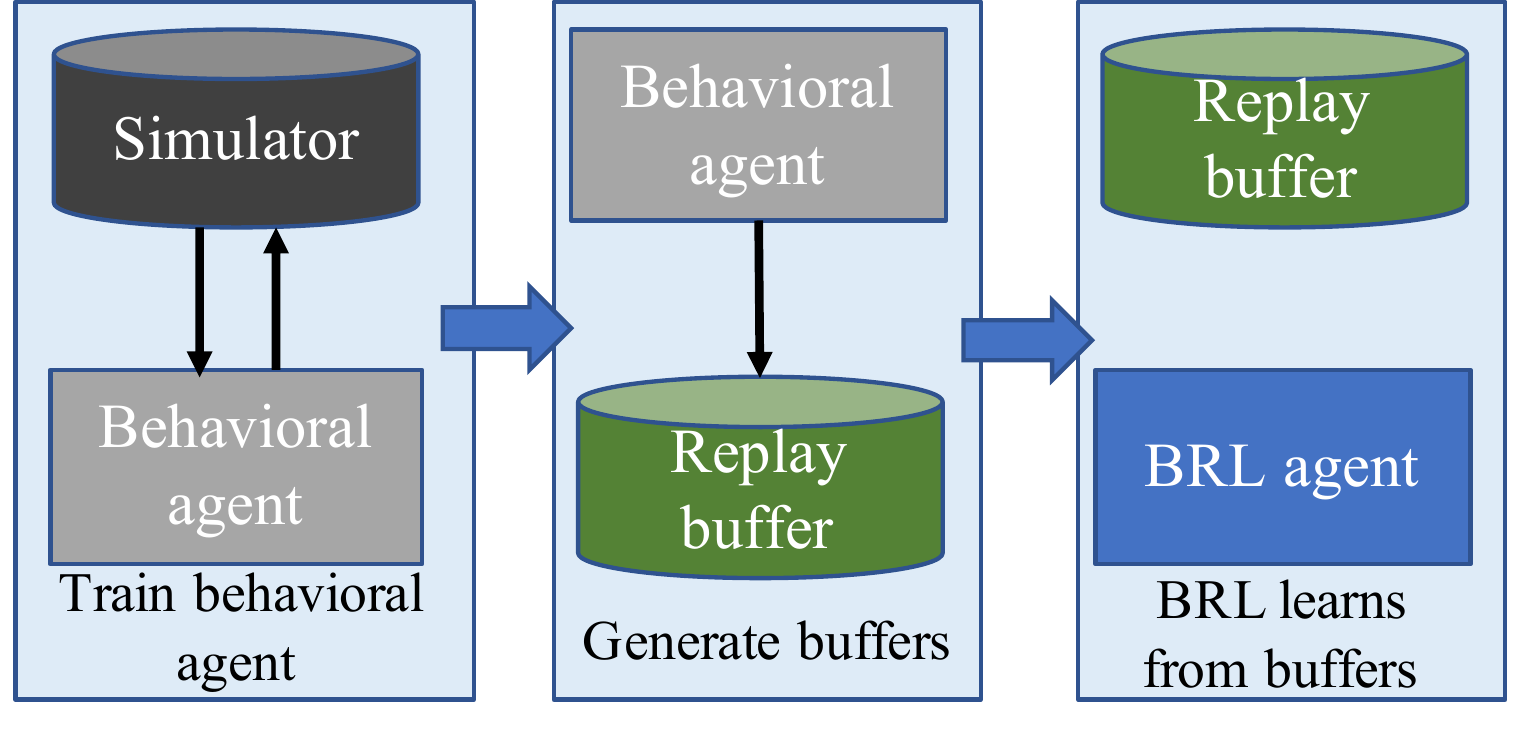}
            \caption{Flow of buffer generation and BRL training}
            \label{fig:flow}
        \end{figure}

            \vspace{-4mm}

  \subsection{Simulated buffers}
    \subsubsection{Data acquisition}
           We adopt Sinergym, an open-source simulation and control framework for training RL 
        agents~\cite{Sinergym}. It is compatible with EnergyPlus models using Python APIs.
            Our approach follows the BRL paradigm. (1) We first train behavioral RL agents for
        $500K$ timesteps and select the one that gives the highest average score as the expert
        agent. 
            Then we run on a 5-zone building, which is a single floor building divided into 5 zones, 1
        interior and 4 exterior with 3 weather types: cool, hot, and mixed in continuous
        settings. We also experiment on two different kinds of response type, deterministic and
        stochastic.
        Then we generate expert buffer with $500K$ transitions as the expert buffer. (2) A medium 
        buffer is generated when the behavioral agent is trained "halfway", which means the 
        evaluation score reaches half of the expert agents' final average scores. (3) We randomly 
        initialize the agent, which samples action from allowed action spaces with uniform distribution to generate buffers. (See Fig.~\ref{fig:flow})
    
        \begin{itemize}
            \item State: Site outdoor air dry bulb temperature, site outdoor air relative humidity,
                    site wind speed, site wind direction, site diffuse solar radiation rate per area,
                    site direct solar radiation rate per area, zone thermostat heating setpoint
                    temperature, zone thermostat cooling setpoint temperature, zone air temperature,
                    zone thermal comfort mean radiant temperature, zone air relative humidity,
                    zone thermal comfort clothing value, zone thermal comfort Fanger model PPD,
                    zone people occupant count, people air temperature, facility total HVAC electricity
                    demand rate, current day, current month, and current hour.
            \item Action: Heating setpoint and cooling setpoint in continuous settings.
            \item Reward: We follow the default linear reward settings, it considers the energy consumption and the absolute difference to temperature comfort. 
        \end{itemize}

    \subsubsection{BRL benchmarks}
        With various qualities of buffers, we compare several most representative benchmarks in the BRL literature and summarize the average scores and standard deviation in the last 5 evaluations across 3 random seed runs (see Table~\ref{table:BRL_scores}). The scores of random policy is normalized to 0 and expert policy is normalized to 100.
 
         \begin{itemize}
            \item TD3+BC: An offline version of TD3, it simply adds a behavior cloning term to regularize actor
            policy towards behavioral policy~\cite{TD3+BC} combined with mini-batch Q-values and buffer states
            normalization for stability improvement. 
            \item CQL: Conservative Q-learning~\cite{CQL}, derived from SAC, learns a lower-bound estimates of
            the value function, by regularizing the Q-values during training.
            \item BC: Behavior cloning, we train a VAE to reconstruct action given state. It simply imitate the behavioral agent without reward signals.
        \end{itemize}
 
\begin{figure*}[ht]
                \begin{center}
                     \begin{subfigure}[t]{0.33\textwidth}
                         \includegraphics[width=\linewidth]{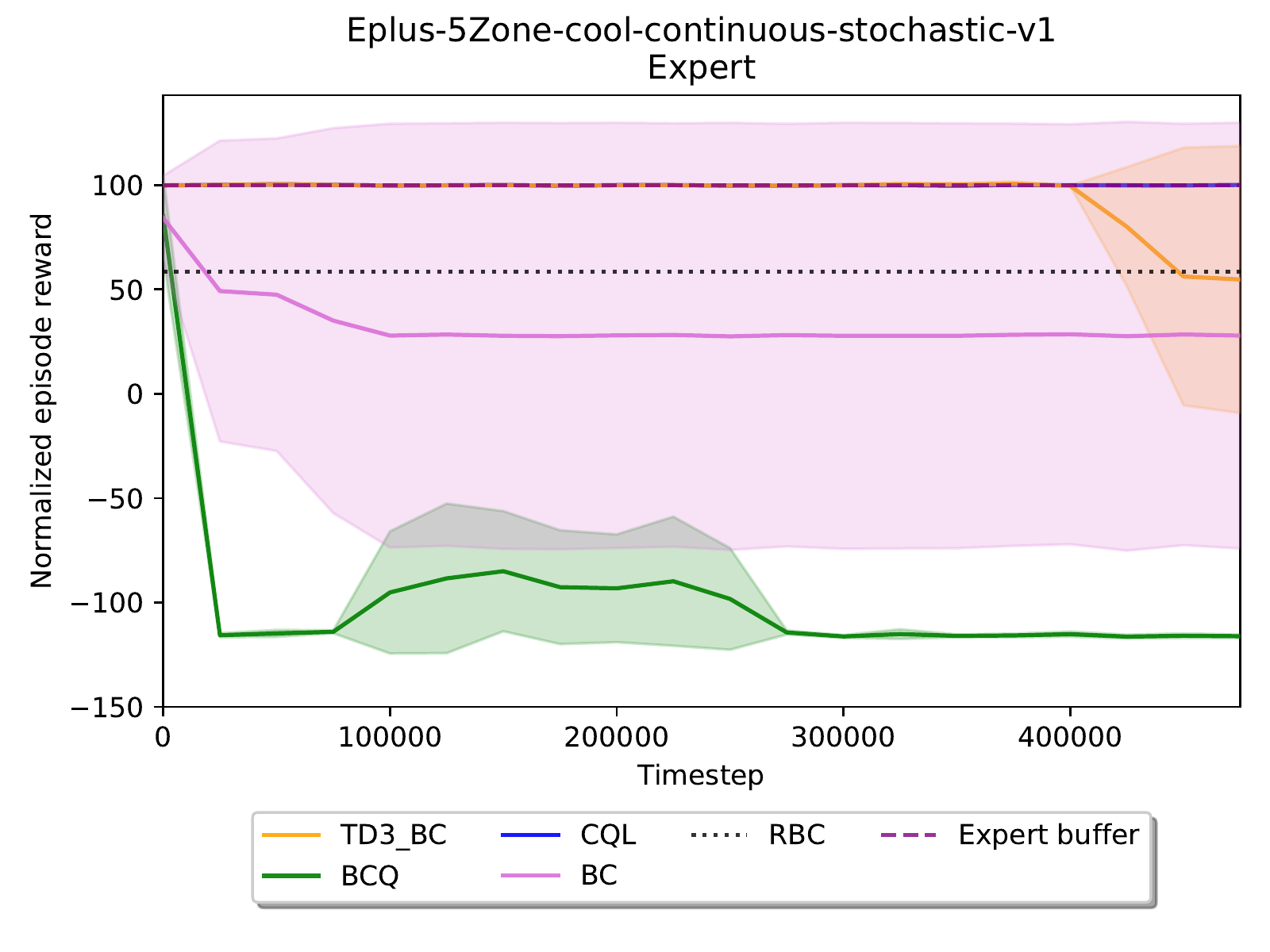}
                     \end{subfigure}
                     \begin{subfigure}[t]{0.33\textwidth}
                         \includegraphics[width=\linewidth]{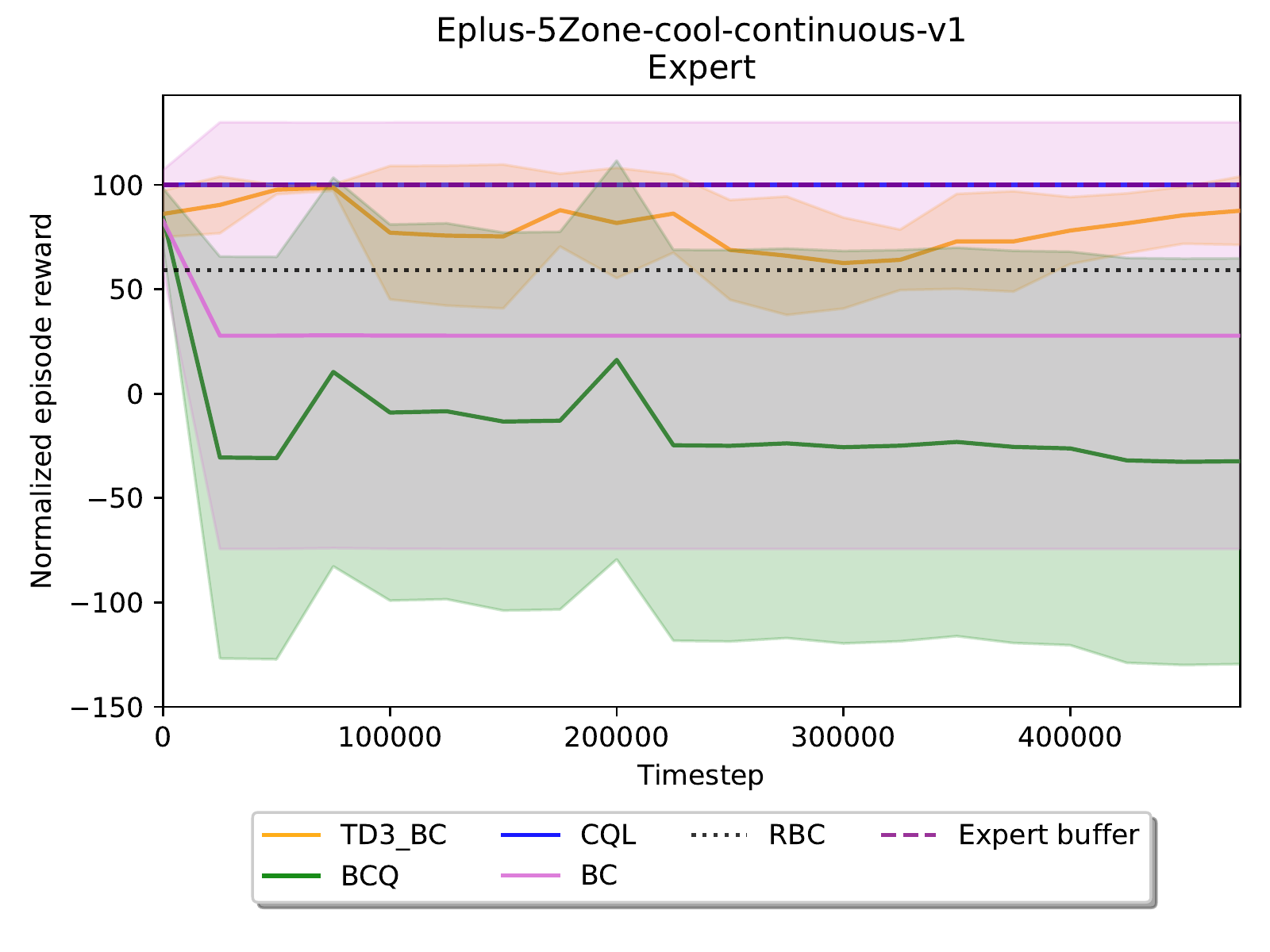}
                     \end{subfigure}
                     \begin{subfigure}[t]{0.33\textwidth}
                         \includegraphics[width=\linewidth]{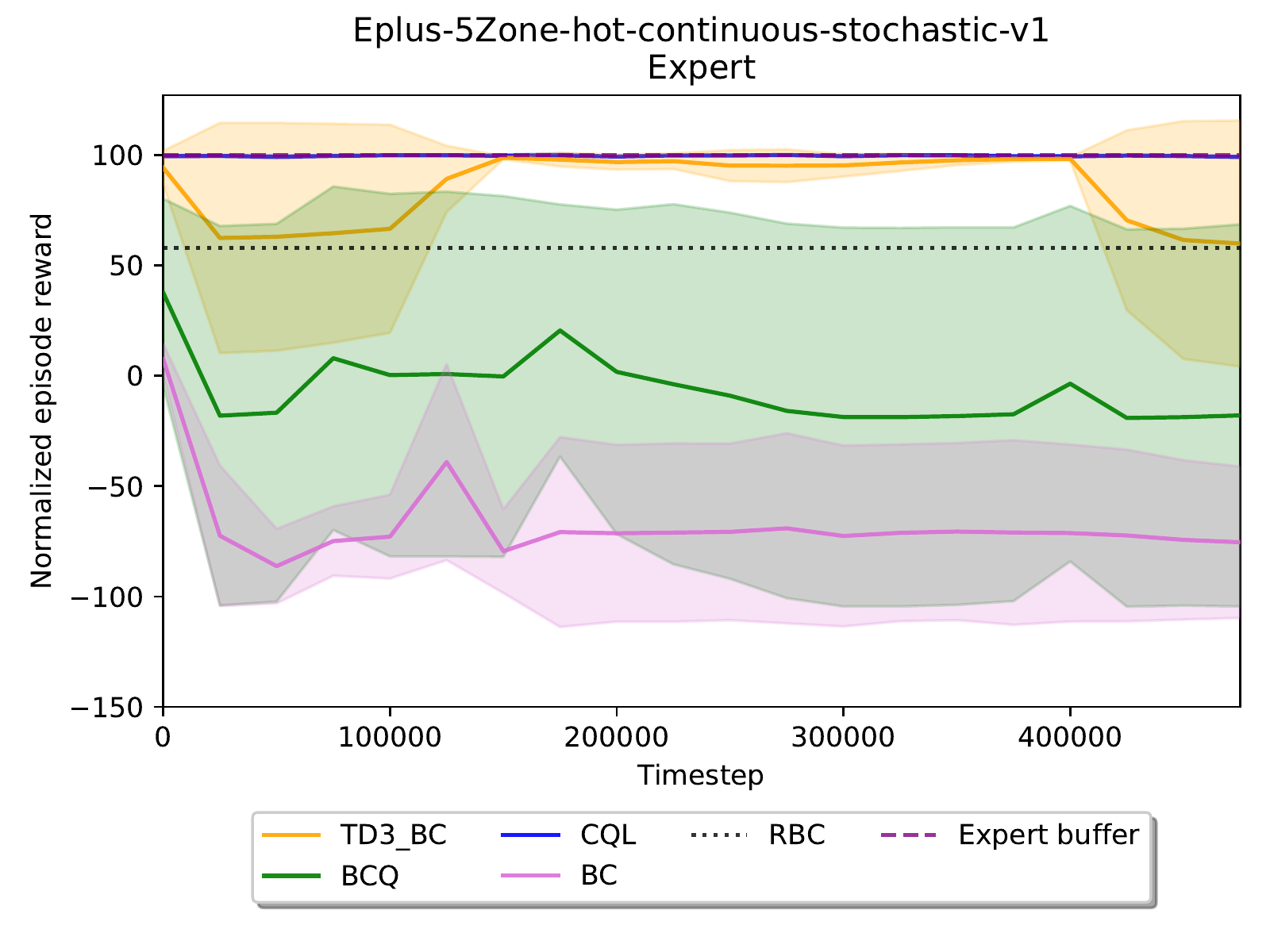}
                     \end{subfigure}
                     \begin{subfigure}[b]{0.33\textwidth}
                         \includegraphics[width=\linewidth]{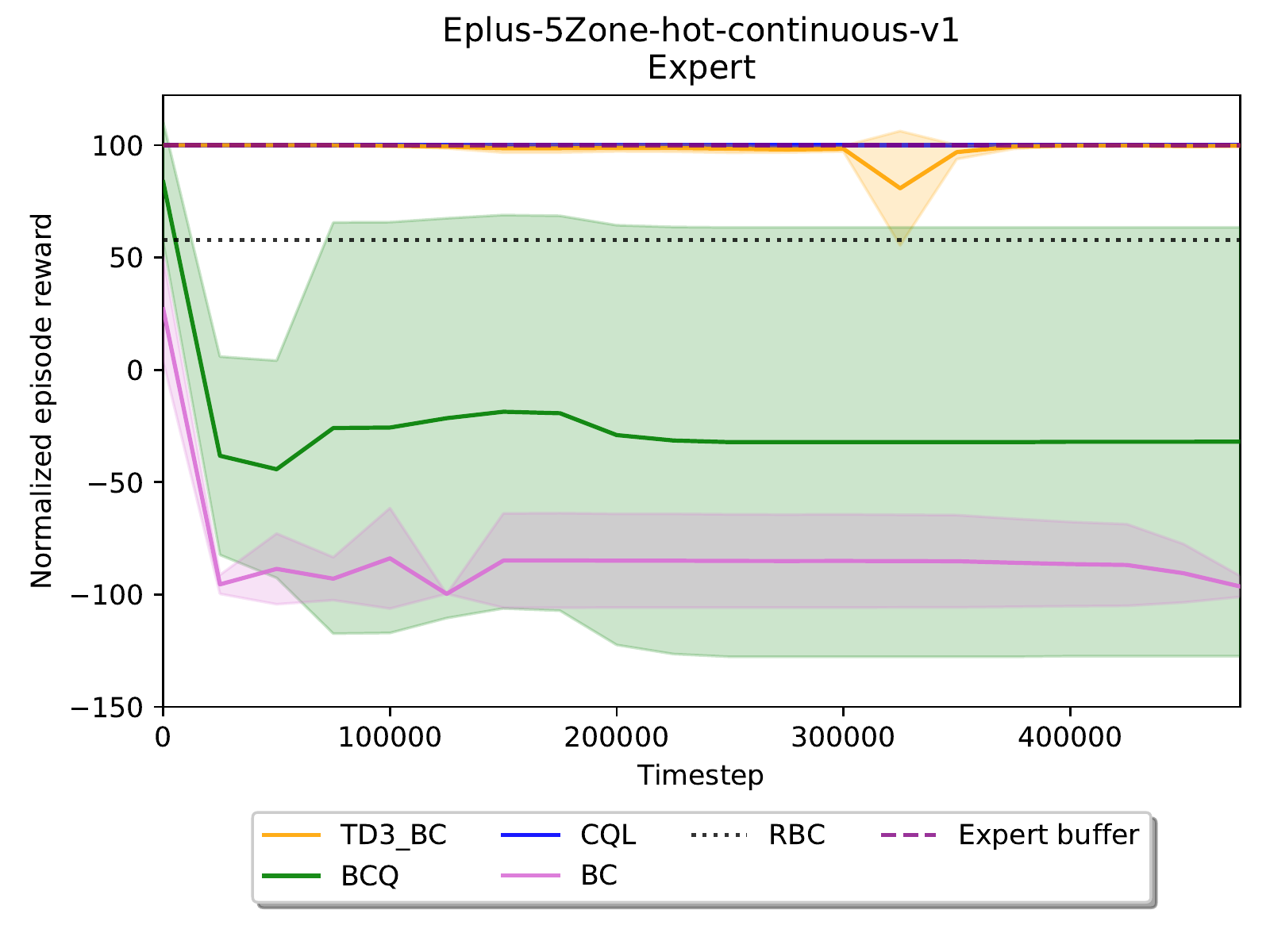}
                     \end{subfigure}
                     \begin{subfigure}[b]{0.33\textwidth}
                         \includegraphics[width=\linewidth]{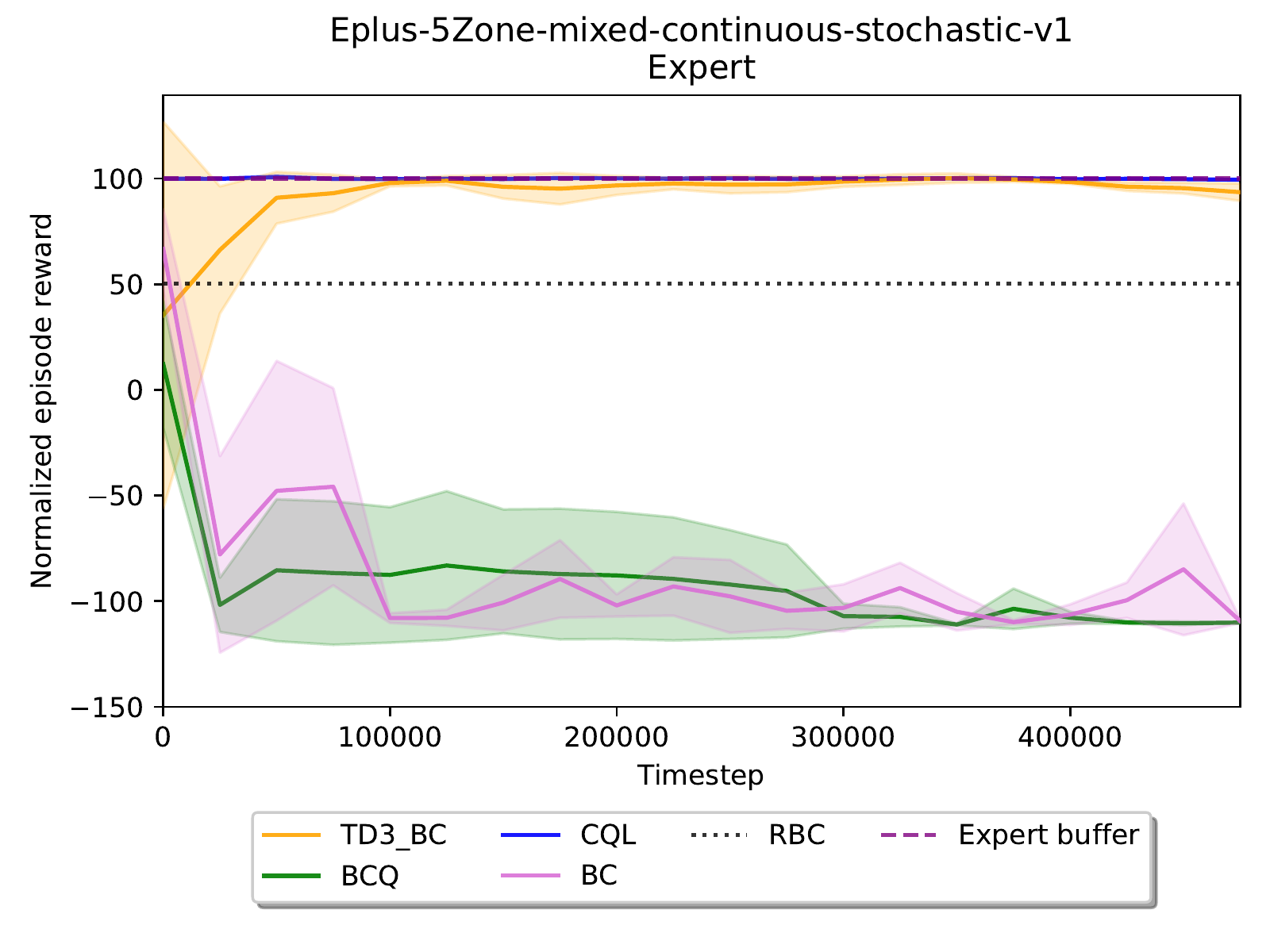}
                     \end{subfigure}
                     \begin{subfigure}[b]{0.33\textwidth}
                        \includegraphics[width=\linewidth]{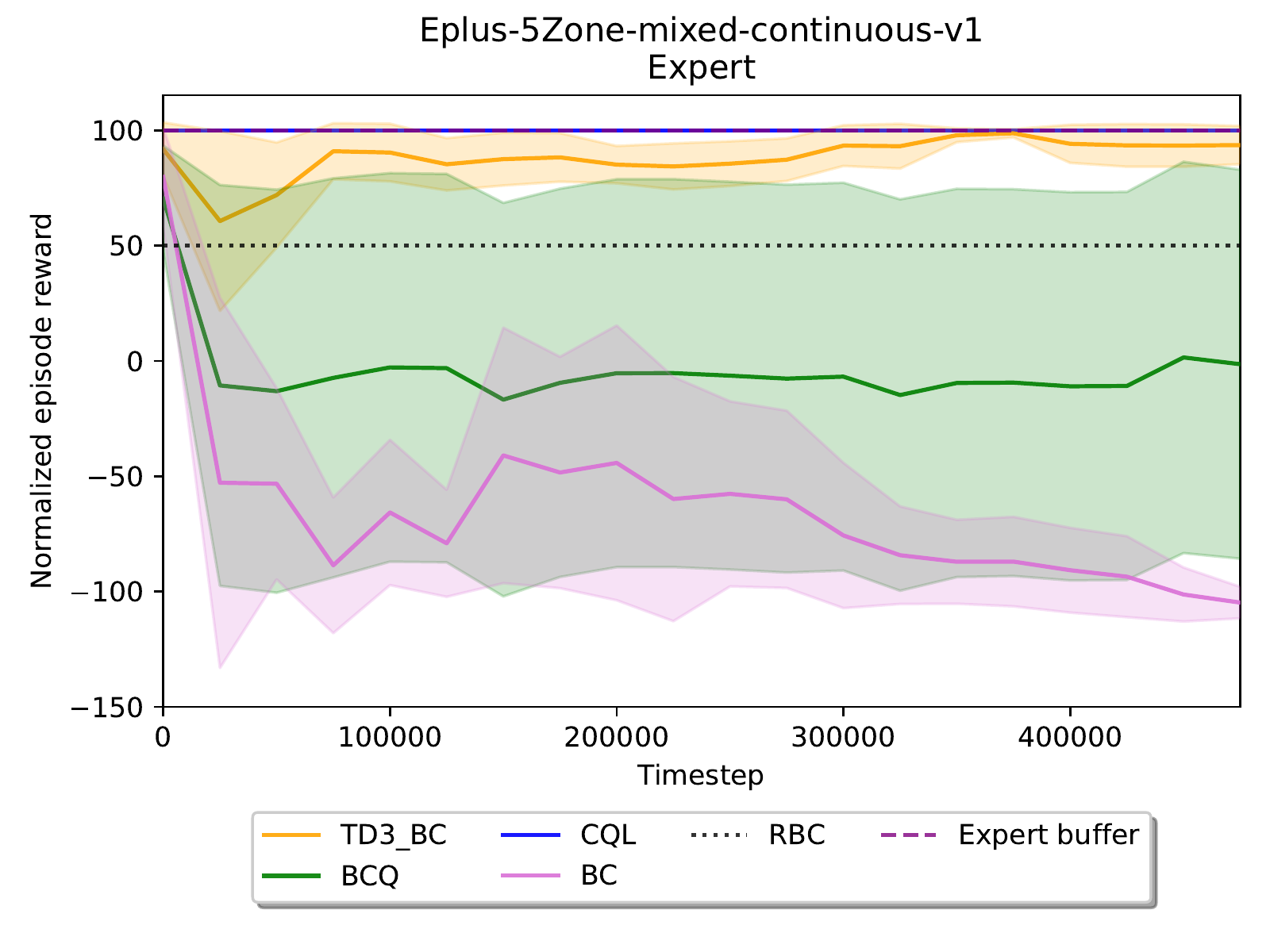}
                     \end{subfigure}
                \end{center}
                \caption{Learning curves of BRL models that learn from expert buffers. Solid line shows the averaged value across three random seeds per algorithm, and the half-transparent region indicates the range with one standard deviation.}
                \label{fig:BRL_expert}
            \end{figure*}

        We train each algorithm for $500K$ timesteps. For every $25K$ timesteps of training we evaluate the models for one episode. As an example, we illustrate BRL learning curves with expert buffers in Fig.~\ref{fig:BRL_expert}.

            \begin{table*}[ht]
            \scriptsize
            \begin{center}
            \begin{tabular}{cccccc}
            \hline
            {Environment} & {Buffer}  & {TD3+BC} & {CQL} & {BCQ} & {BC}
            \\ \hline
            hot-deterministic & Expert & \textbf{99.72±0.1}  & \textbf{100.00±0.00} &  -32.02±0.07 & -89.2±3.95\\
            hot-deterministic & Medium  & -49.59±8.19  & \textbf{67.65±17.06} & 13.41±16.59 & -12.55±7.27 \\
            hot-deterministic & Random   & -45.73±15.13 &-23.19±4.52 & \textbf{69.21±18.52} & -26.74±15.91\\
            \hline
            mixed-deterministic & Expert   & \textbf{94.67±2.04}   &\textbf{100.00±0.00}  & -6.22±5.24 & -95.46±6.6\\
            mixed-deterministic & Medium  & 36.23±4.31   &37.36±19.31  & \textbf{64.46±0.65} & -103.4±2.12\\
            mixed-deterministic & Random  & -13.72±22.25 &-23.46±20.33 & -65.30±20.40 & -27.82±11.79\\
            \hline 
            cool-deterministic & Expert   & 81.11±5.24   &\textbf{100.00±0.00}  & -29.75±3.18 & 27.76±0\\
            cool-deterministic & Medium   & -49.97±0.00  &55.44±6.46 & \textbf{70.19±17.06} & 10.48±22.11\\
            cool-deterministic & Random   & -58.40±3.21  &12.99±2.28 & \textbf{27.77±31.39}  & 8.62±41.97\\
            \hline
            hot-stochastic & Expert     & 77.69±17.18  &\textbf{99.49±0.20} & -15.35±5.92  & -72.86±1.73\\
            hot-stochastic & Medium    & -14.85±0.00  &\textbf{39.93±2.64} & -62.21±19.31  & -10.45±12.85\\
            hot-stochastic & Random    & -1.82±2.68   &\textbf{36.65±11.95} & -1.24±14.80  & 31.22±13.51\\
            \hline
            mixed-stochastic & Expert & \textbf{96.61±2.13}   & \textbf{99.77±0.26} & -108.38±2.58  & -102.02±9.32\\
            mixed-stochastic & Medium   & 9.49±0.00    & \textbf{80.13±8.19} & 70.75±6.46 & -107.41±3.41\\
            mixed-stochastic& Random   & 28.02±8.69   & \textbf{94.05±2.08} & -109.47±0.17  & 38.66±24.64\\
            \hline
            cool-stochastic & Expert  & 78.27±20.01  & \textbf{99.97±0.12} & -115.86±0.41 & 28.15±0.35\\
            cool-stochastic & Medium    & 16.09±0.00   & \textbf{81.57±4.31} & -11.55±2.64 & -50.37±2.45\\
            cool-stochastic & Random    & -44.33±16.01 &-97.35±2.09 & -53.92±10.07 & \textbf{25.44±13.42}\\
            \hline
            \multicolumn{2}{c}{Sum}  &339.50±127.23 & \textbf{960.99±101.81}& -295.49±175.47  & -527.93±193.48\\
            \hline
            \end{tabular}
            \end{center}
            \caption{Average normalized score over the final 5 evaluations and 3 random seeds. ± 
            corresponds to standard deviation over the last 5 evaluations across runs.}
            \label{table:BRL_scores}
            \end{table*}

%% file: 4_conclusion_discussion.tex
\section{Conclusion and Future Works}
    \label{sec:discussion_conclusions}
    We open-source our building control datasets for both real buildings and simulation environments as BRL benchmarks. The goal is to encourage building domain experts to explore opportunities in building-BRL research. We provide these datasets for researchers to implement fast prototyping without generating buffers on their own.
    Recently, many building-RL libraries are published~\cite{beobench, energym, citylearn} for the purpose of building RL training without the need to set up thermal simulators beforehand. Our future work is to generate more diverse buffers with various building environments and different weather types for BRL benchmarks.